\documentclass[conference]{IEEEtran}
\IEEEoverridecommandlockouts
\usepackage{cite}
\usepackage{amsmath,amssymb,amsfonts}
\usepackage{algorithmic}
\usepackage{graphicx}
\usepackage{textcomp}
\usepackage{xcolor}
\def\BibTeX{{\rm B\kern-.05em{\sc i\kern-.025em b}\kern-.08em
    T\kern-.1667em\lower.7ex\hbox{E}\kern-.125emX}}
\begin{document}

\title{Target Reaching Behaviour for Unfreezing the Robot in a Semi-Static and  Crowded Environment \\

\thanks{This work has been funded by the European Union's Horizon H2020 Research and Innovation program Crowdbot project under grand agreement No 779942.}
}

\author{\IEEEauthorblockN{Arturo Cruz-Maya}
\IEEEauthorblockA{\textit{AIRC, Software department} \\
\textit{Softbank Robotics Europe}\\
Paris, France \\
arturo.cruzmaya@softbankrobotics.com}

}

\maketitle

\begin{abstract}
Robot navigation in human semi-static and crowded environments can lead to the freezing problem, where the robot can not move due to the presence of humans standing on its path and no other path is available. Classical approaches of robot navigation do not provide a solution for this problem. In such situations, the robot could interact with the humans in order to clear its path instead of considering them as unanimated obstacles. In this work, we propose a robot behavior for a wheeled humanoid robot that complains with social norms for clearing its path when the robot is frozen due to the presence of humans. The behavior consists of two modules: 1) A detection module, which make use of the Yolo v3 algorithm trained to detect human hands and human arms. 2) A gesture module, which make use of a policy trained in simulation using the Proximal Policy Optimization algorithm. Orchestration of the two models is done using the ROS framework.
\end{abstract}


\section{Introduction}

In  the last  years,  an increasing amount of robots  have  been  deployed  in  public  environments  making  evident  the  need  of human-aware navigation capabilities. Environments where robots are capable of navigating such as hospitals, hotels, restaurants, train stations or airports have been subject of research studies \cite{joosse2017guide}\cite{sasaki2017long}\cite{mishraa2018issues}.

Classical  robot  navigation  approaches tend  to  focus on efficiency during the path planning and execution; variables such as path length, clearance and smoothness are used to quantitatively evaluate performance \cite{tzafestas2018mobile}. Nevertheless, robots using such approaches tend to disturb the crowd flows and to move in unexpected manners. Tackling these problems, more recent works have  have proposed different strategies for socially acceptable navigation for robots in complex environments where human presence and activities are unpredictable. These works, can be classified into two groups: model-based and learning-based methods.

The major exponents of model-based methods rely on social psychology and cognitive sciences to generate human-like paths for robot navigation. One of the most relevant approaches is the Social Force Model (SFM) \cite{helbing1995social}, which proposed to model pedestrian’s behaviour whose motion is influenced by other pedestrians by means of repulsive forces. Many studies have implemented this method but also produced several variations \cite{zanlungo2011social}, later applied to real world environment navigation where robots can avoid or go along with people \cite{ferrer2013robot}. However, these  works  show  limitations  such  as  the  need  for  parameter  calibration  in  different  robots  or  the  need  of additional sensors for pedestrian tracking. Also, model-based methods are based on geometric relations, but it is still unclear if pedestrians always follow such models.

\begin{figure}[t]
\centerline{\includegraphics[height=42mm]{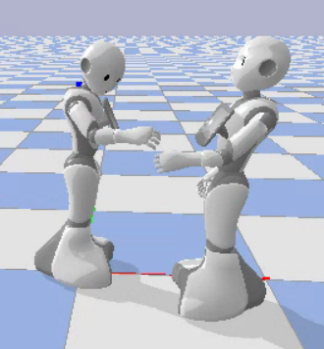}
\includegraphics[height=42mm]{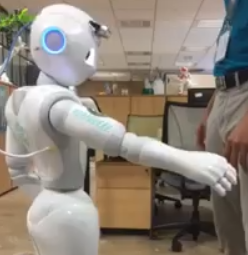}}
\caption{Left: Target reaching behavior on simulation. Right: Target reaching behavior on the real robot}
\label{figTargetReaching}
\end{figure}

In contrast, learning-based methods use policies for defining human-like behaviours, which are usually learnt from human demonstrations by matching feature statistics about pedestrians. These methods apply machine learning techniques such as Inverse Reinforcement Learning (IRL) to model the factors that motivate people’s actions  instead  of  the  actions  themselves.  An  experimental  comparison  of  features  and  learning  algorithms based  on  IRL  is  presented  in  \cite{vasquez2014inverse},  where  the  authors  conclude  that  it  is  more  effective  to  invest  effort  on designing  features  rather  than  on  learning  algorithms.  More  recent  approaches  include  the  use  of  deep reinforcement  learning  in order  to  set  restrictions  \cite{chen2017socially}  (i.e.  passing  at  the  left  side  of  the  people)  instead of learning the features that describe human paths.

Nevertheless, few efforts have been made concerning the problem of labelling humans as an interactive agents when blocking the robot motion trajectory. Current state of the art navigation planners will either propose an alternative path (if possible) or freeze the  motion  until  the  path  is  free.

Our work is part of the CrowdBot European project \footnote{http://crowdbot.eu/}. This project articulates ethical questions with technical problems in the goal of enabling robots, autonomous ones or semi-autonomous ones to navigate in crowded environments. Navigating through human crowds is a tough challenge for a robot. Crowds can cause severe sensor occlusions and often don’t leave much free space for the robot to move in, leading to what’s known as the  "freezing robot problem". Furthermore, we contemplate semi-static scenarios where the robot can not pass unless it interact with the humans blocking its path. In a previous work \cite{cruz2019enabling} we presented a robot behavior for interactive navigation used in deadlock situations. In such work the robot used speech with the purpose of asking for permission to pass when its path was blocked by humans. Nevertheless, there exist situations where a physical contact is needed when a robot is navigating, as is the case of crowded and noisy environments (i.e. a cocktail party), where people standing and talking are unaware of the intentions of the robot.  

The goal of this work is to provide a wheeled humanoid robot with a behavior that is capable of unfreezing the robot in deadlock situations in semi-static, crowded and loud environments. One of the constraints on the design of this behavior was that the modality should be other than audio, as a the voice of the robot could be not heard in a loud environment.  Also, as the robot used in this work is the humanoid Pepper robot, we take advantage of the physical modality. This modality should respect social norms such as touching only body parts that are socially allowed. A  study on  the meaning  of  touch  [18]  showed  that  the  hands  and  shoulders  are  considered  regions  that  serve to direct the recipient's attention, also most of the touches were initiated by hand alone. 

A robot behavior that could be used in the context of this work is the one of touching the shoulder of a person from the back, unfortunately the robot used in our work is too small to perform such behaviour. Then, we decided that the behavior should consist of moving one arm of the robot until reaching the hand of the person blocking the path. Furthermore, in order to increase the safety of the robot behavior, it should track the target with its head while performing the movement.

\section{Related Work}

\subsection{Unfreezing Robot Problem}

Navigation methods described in the introduction, usually focus on the motion of the robot. The work presented in \cite{trautman2010unfreezing} presents an approach called "Interactive Gaussian Processes" (IGP) based on the assumption that agents solve the freezing problem by engaging in a cooperative strategy to create feasible trajectories, such approach should work for agents in motion.
A more recent approach for unfreezing the robot in crowded environments, is presented in \cite{fan2019getting} which consist of a  recovery behavior in case of SLAM failure, this behavior used a reinforcement learning based local navigation policy to find a set of recovery positions with rich features which allow the robot to continue its navigation.

Nevertheless, there are situations where more complex social behaviours are needed in order to deal with the freezing robot problem.  Only a few research studies have focus on interactive navigation using robot capabilities.
An approach for management of deadlock situations at narrow passages is presented in \cite{trinh2015go}, where the robot lets the conflicting person pass and waits in a non-disturbing waiting position. Other work  \cite{vega2018planning}  proposes  an interactive navigation  in two simulated scenarios:  In  the  first  one,  the  robot  asks  for collaboration to enter a room. In the second one, the robot asks for permission to navigate between two people who are talking. In another work \cite{kamezaki2019preliminary}, the authors of propose the use of multi-modal rule-based system for interactive navigation, the interaction modals included visual, acoustics, and haptics, where the possible actions of the robot for clearing its path were: robot movement, speech, arm contraction, passive touch, and notifying touch, results showed that their system solved freezing problems, provided a safe and efficient trajectory while improving humans perception. 

\subsection{Target Reaching Behavior}

The choice of using a target reaching behavior for solving the unfreezing the robot problem relates our work with several works on the target reaching problem, which is one of the most important problems in robotics, as it is needed for tasks such as manipulation, grasping and object interaction. 

The problem of the movement of a kinematic chain given a desired position of the end effector is a well-known problem in classical robotics known as Inverse Kinematics (IK) where the most common approach is to use a Jacobian matrix to find a linear approximation \cite{buss2004introduction}. There  are  many  public  software  libraries for  solving motion planning for robotics. However, IK solvers do not ensure a solution and they could require variable time to find one. Approaches using neural networks has been proposed for stereo-head robot arm coordination\cite{guerrero1999neural} allowing the robot to reach a given a spatial target using different joints. Also, using genetic algorithms with neural networks has allowed robotic arms to reach target at random target locations while avoiding obstacles \cite{moriarty1996evolving}
Novel approaches for solving the target reaching problem include the use of Spiking Neural Networks (SNN) \cite{maass1997networks}, which shows promising results but are still far for being a practical solution.

Current approaches in the state of the art for generating robot motions use Deep Reinforcement Learning (DRL) methods, i.e. \cite{haarnoja2018learning} a quadruped robot learns to walk using such approach. Controllers based on DRL can be very computationally efficient at runtime, they provide a control policy learning system, but they do suffer from extremely slow training times. Because of this, the common way to train DRL in robotics is through simulation, an example of this is the work of \cite{peng2020learning} where the authors teach a simulated robot to move like an animal, then the learned policy is adapted to a real robot. An  empirical  evaluation  of  off-policy  DRL algorithms  on  vision-based  robotic  grasping  tasks in a simulation environment is presented in \cite{quillen2018deep}. 

DRL has been used before for teaching the Pepper robot to interact with humans, in \cite{qureshi2016robot} the authors propose the use of a dual stream convolutional neural network, the actions of the robot include waving hand, waiting, looking towards the human, and handshake.

In this work, we use a DRL method to perform a target reaching behavior on the Pepper robot given a dynamic spatial target.

\section{Proposed System}
\begin{figure}[t]
\centerline{\includegraphics[width=90mm]{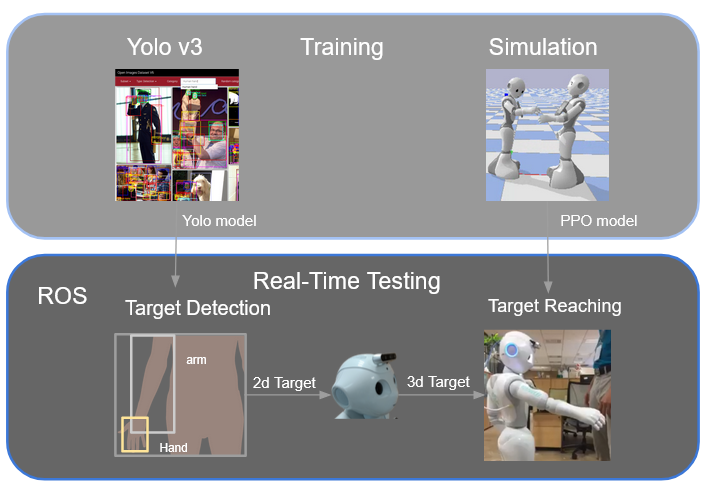}}
\caption{Architecture of the proposed system}
\label{figSystem}
\end{figure}

The proposed Target Reaching Behavior is composed of two modules that are coordinated using the Robotic Operative System (ROS) framework, see Fig.\ref{figSystem}. The first module, in charge of the perception of the robot. This module consist of a target detector using the Yolo v3 algorithm trained on the Open Image Dataset. The target detection module outputs a 3D target position using a RealSense D435 device set on the head of the robot.
The target detector module is described in section IV. The second module, in charge of the actions of the robot, consist of a DRL method using the Proximal Policy Optimization algorithm trained on a physical simulator. The target reaching module is presented in section V.

\section{Target Detection}

The target of our system for reaching by the robot are human hands and human arms that will be located at a close range distance. 

There exist solutions in the state of the art of hand detection that are readily available. One of these  systems  is  OpenPose [19],  which can not only  detect hands,  but  also  the  whole  human  skeleton. Nevertheless,  OpenPose  was  proved  to  be  unsuitable  for  our  needs, this is because of the short  distance between the robot and the person in front of it in our scenario. Also, because of the specifications of the RGB-D sensor and the size of the robot, we need a system capable of detecting hands at a distance between 0.2 m. and 0.8 m.

Due to the high success of deep neural networks in computer vision tasks, YOLOv3 \cite{redmon2018yolov3} was chosen as the object detector to be used, which is one of the state-of-the-art deep learning algorithms that perform real-time object detection.

\subsection{Dataset}

The system  uses,  as  other  deep  learning  methods, a supervised  learning approach for which  it  needs  to  be trained on a database of annotated images. For this reason, the Open Images Dataset(OID)\cite{kuznetsova2018open} was selected. This database provides a large collection of labelled images and their bounding boxes. This dataset was chosen because the images are very diverse and often contain complex scenes with several objects,allowing the model to learn more complex patterns.
For training the target detector, two classes were selected: human arm, and human hand. For each class, 10,000 images  were  downloaded.  The  arm  class  allows  a  secondary  target  in  case  the  hands  of  the  person  are  not visible or are not recognized.

\subsection{YOLO algorithm}
The version of YOLO used in this work is the implementation of PyTorch-YOLOv3\footnote{https://github.com/eriklindernoren/PyTorch-YOLOv3\#pytorch-yolov3}.
The images labels were pre-processed in order to meet the requirements of YOLOv3. The network was trained from scratch with the images downloaded from the Open Images Dataset.

\subsection{YOLO Training}
The model was trained for 400 Epochs of 2400 steps each epoch, the mean average precision (mAP) on the testing  dataset  was around 0.24\%  and  a  loss  value  of  0.67,  while  it  still  has  room  for  improvement  it  is acceptable for our needs. 

\begin{figure}[!t]
\centerline{
\includegraphics[width=45mm]{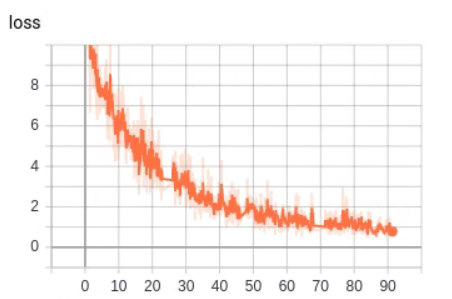}
\includegraphics[width=45mm]{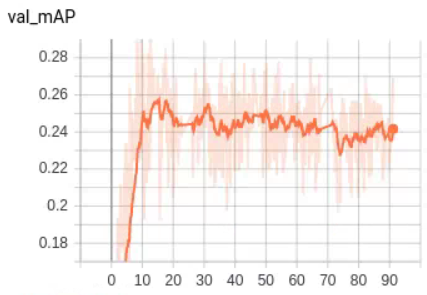}}
\caption{Loss and mAP of the neural network showing the relative time for each step using TensorBoard}
\label{figYoloLoss}
\end{figure}

The training took about 30 hours on a PC composed of a 1080 Tegra Ti GPU. Further improvement on the mAP would require more and more time and it will increase only a small fraction of the current result. The loss value of the neural network is presented in Fig.\ref{figYoloLoss}  showing the relative time for each step, as the training was done in separate times each 100 epochs, it can be seen that the log after 100 epochs had descended to 2.82, after 200 epochs it was 1.14, after 300 epochs it was 0.83 and the final value after 400 epochs was 0.67. In the validation set, the arms present a better mAP (0.33) than the hands (0.27).

\begin{figure}[!b]
\centerline{\includegraphics[height=40mm]{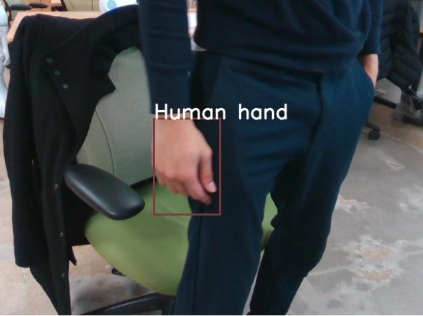}}
\caption{Output of the trained model using images streamed from the Realsense D435 camera set on the head of the robot.}
\label{figYoloReal}
\end{figure}

The model was also tested with images streamed from the RealSense depth camera D435 set on the robot in order to verify that it fulfils the needs of this module. An example is shown in Fig.\ref{figYoloReal}, where it can be seen that the model is able to detect human hands at a close distance.

\subsection{Model Inference}

The  inference  is  done using  the  PyTorch  framework inside a ROS package  in  order  to  facilitate  the communication with the other modules of the  project. Depending  on  the  availability  of  CUDA  the  implementation  of  the  model  uses GPU  or  CPU  to  make  the inference. The  detections  are  re-scaled  to  the original size of the image, discarding too small (1:10) or too big images (8:10) using empirically obtained thresholds.

In order to speed up the conversion of the depth points in the region of interest to a single 3D position, the 3D target is calculated taking the depth  information  of n random  pixels (pre-defined to 10) from  the  center  of  the bounding  box and computing its mean. Afterwards, the ROS tf2 package is used to translate the hand position to the base of the robot in order to perform the hand gesture. Finally, the target is published to a ROS topic if it is a new detection or if the previous published point is a range of 0.2m.

\section{Target Reaching Behavior}

In this section, we define our formulation for the target reaching behavior. Then we describe the training of the method in simulation. Finally we show the results in the real robot.

\subsection{Problem Statement}

The goal of this module is to generate a robot movement for allowing it to reach a given 3D position target with its hand. The scenario where the robot gesture will be used, which is a Human-Robot Interaction, imposes some restrictions such as the movement of the robot needs to be fast enough to follow changes in the position of the target (hand or arm), also as the size of the robot is small compared to an adult person, using the hip joint to reach a point with the hand would be useful as this increase the action space of the robot. Currently, the NAOqi framework has a module for pointing with the arm to a given 3D position. Nevertheless, it has some limitations like the movement does not include the hip and the head of the robot and it is done in a blocking call, which means that the robot will execute following orders until the movement is done. This is a disadvantage for having a tracking with the head at the same time. Instead, having a head tracking the target and maintaining it in the field of view of the robot could be of aid for assuring a reactive gesture.

\subsection{Problem Formulation}

Policy gradients methods are one of the major pieces in recent improvements in deep learning for control. Based on this, we formulate the problem of finding a trajectory of joint positions for a kinematic chain of a robot as a partially observable Markov decision process as a tuple 
\begin{math} M\{S, O, A, T, R\}\end{math}
where $S$ is a set of states of the agent, $O$ is a set of observations, $A$ is a set of Actions, $T$ is a state transition probability function, where \begin{math}T(s_{t+1}  s_{t}, a_{t})\end{math}, and $R$ is a reward function $R$ = $S$ $X$ $A$ $\rightarrow$ $R$.

We chose the Proximal Policy Optimization (PPO) algorithm \cite{schulman2017proximal} for learning the policy for our target reaching behaviour. PPO seeks to improve previous policy gradient methods, while using only first-order optimization. Also, PPO uses an objective with clipped probability ratios, which forms a lower bound of the performance of the policy. To optimize policies, it is alternated between sampling data from the policy and performing several epochs of optimization on the sampled data.

We describe the observations, actions, and reward as follows:

\subsubsection{Observations}
The observations $O$ are represented as [$joint\_angles$, $robot\_hand$, $target$, $\theta$ $head\_t$, $\theta$ $head\_w$]
where $joint\_angles$ are the the angle in radians of the joints $j$ of the kinematic chain, 
$robot\_hand$ is the 3D position $[x,y,z]$ of the right hand of the training robot,
$target$ is the 3D target position $[x,y,z]$,
$\theta$ $head\_t$ is the  direction vector of the head of the training robot $[x,y,z]$, 
and $\theta$ $head\_w$ is the direction vector of the position of the head of the training robot towards the position of the hand of the second robot $[x,y,z]$. 
All observations are normalized between $[-1, 1]$. 

\subsubsection{Actions}
The actions $A$ are represented as [$joint\_velocities$] the normalized velocities of the joints. The velocities were also limited to $0.75$ for the head joints and to $1.0$ for the rest of the joints.

\subsubsection{Rewards}

We use a combination of two rewards, the first one is related to the movement of the arm of the robot, and the second one is related to the movement of the head.

The arm reward is defined as:

\begin{equation}
    a\_r = exp(-1 \| robot\_hand - target \|)
\end{equation}

The head reward is defined as:

\begin{equation}
    h\_r = exp(-1 \| \theta head\_t - \theta head\_w \|)
\end{equation}

The rewards were combined as follows:
\begin{equation}
    r = w_1 * a\_r + w_2 * h\_r
\end{equation}

Where $w_1=0.75$ and $w_2=0.25$ are weights empirically obtained, a greater weight for the head reward would not allow the learning of the rest of the joints which need a more complex behavior.

\subsection{Training Environment}
In this work, OpenAI Baselines\footnote{https://github.com/openai/baselines} are used, it is a set of implementations of RL algorithms of the state of the art, more specifically a fork of Baselines called Stable Baselines\footnote{https://stable-baselines.readthedocs.io/en/master/index.html} [28] is used, which provides unified structure and documentation for the algorithms. 

In order to train the model, we used a physical simulator called qiBullet \cite{busy2019qibullet} which is a Bullet-based simulator for the Pepper and NAO robots. The qiBullet simulator inherits the cross-platform properties of the PyBullet Python module and Bullet physics engine. Also, PyBullet is a fast and easy to use Python module for robotics simulation and machine learning, with a focus on sim-to-real transfer. 

The simulated environment used in this work involves two robots, one of them is controlled by the PPO algorithm while the other is set at random positions. The goal is to train a policy able to perform a hand gesture that touches the hand of the other robot regardless of its position, while tracking it with the head and the right hand of the robot. The second robot is used because of simplicity. The simulation runs at the default rate of 1/240 seconds using the method stepSimulation().

\subsubsection{Robot model}
The model of the simulated Pepper robot is given by the qiBullet simulator. The joints for training the target reaching behavior are: hip pitch, right shoulder pitch, right shoulder roll, right shoulder elbow, head yaw, and head pitch. Other joints were excluded even if they could improve the action space of the robot, because it would lead to loss of naturalness of the motions.

\subsubsection{Target}
The target is given by the 3D position of the left hand of a second robot having the world as reference frame. The position of the second robot and its left hand are generated randomly at each episode inside a region in front of the robot to be trained. The x coordinate is set in a range of $[0.65, 0.85]$ and the $y$ coordinate  is set between $[-0.3, 1]$. The left shoulder pitch is set in a range of $[0.4 – 0.8]$. This range gives a target space that contains several reachable targets and some that are not. 
The position of the second robot is intentionally set at the right side of the training robot in order to facilitate the learning as the physical design of the robot poses further restrictions to the reaching space. Furthermore, an offset of $5$ cm in the $y$ axe was added to the target, so the robot would not try to reach the joint position inside the hand of the other robot.

\subsubsection{Training}
\begin{figure}[!t]
\centerline{\includegraphics[width=0.48\textwidth]{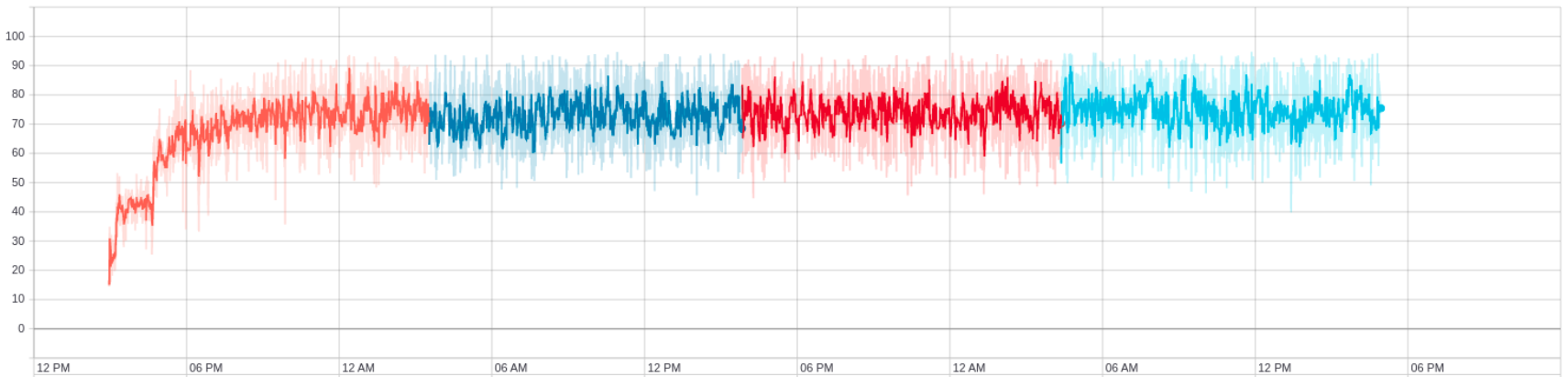}}
\caption{Discounted reward of the teaching reaching behavior and head tracking using the PPO algorithm}
\label{figDiscountedReward}
\end{figure}

Training the robot in simulation, the combined reward increased in the first 10 million steps, after that it decreased and only got similar results after more than 30 million steps. The stop condition to restart the episodes was when the arm of the robot made a collision with itself, and after 5 seconds of the beginning of the episode. The model was saved each 10 million steps. The discounted reward of the training is shown in Fig.\ref{figDiscountedReward}, where each color represents 10 million episodes

\begin{figure}[!b]
\centerline{
\includegraphics[height=22mm]{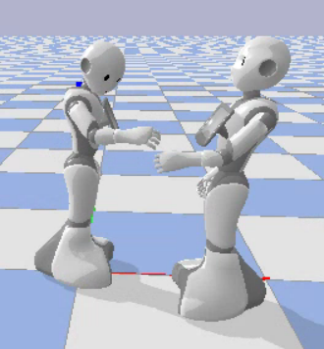}
\includegraphics[height=22mm]{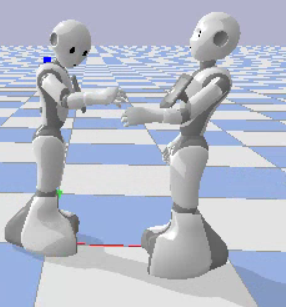}
\includegraphics[height=22mm]{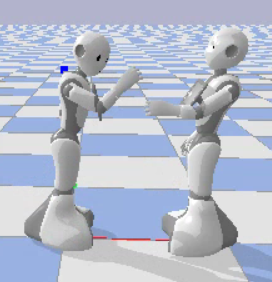}
\includegraphics[height=22mm]{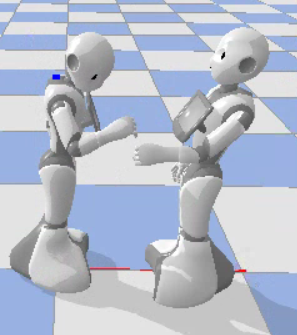}
}
\caption{Side view of the target reaching behavior with 4 targets at different positions using the learned policy }
\label{figSideView}
\end{figure}

\begin{figure}[htbp]
\centerline{
\includegraphics[height=22mm]{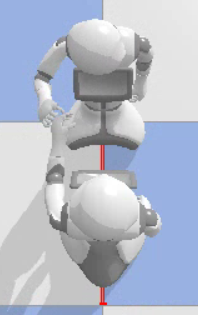}
\includegraphics[height=22mm]{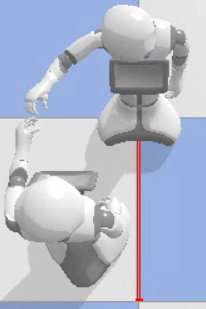}
\includegraphics[height=22mm]{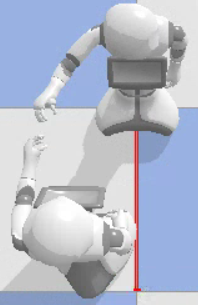}
\includegraphics[height=22mm]{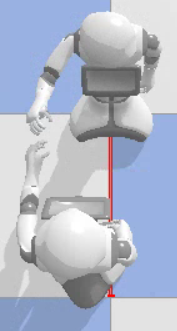}
}
\caption{Top view of the target reaching behavior with 4 targets at different positions using the learned policy}
\label{figTopView}
\end{figure}

The robot achieved to reach the target at different positions that were randomly generated, also it was able to adapt its movement to dynamic targets. Different robot motions for different target positions are shown in Fig.\ref{figSideView} where the motion can be seen from a side view, and Fig.\ref{figTopView} where the motion can be observed from a top view. 

\section{Real Robot Implementation}
This section present the target reaching module implemented in ROS for using it in the real robot.

The Target reaching ROS module is composed of 3 nodes: 
\begin{itemize}
    \item Target frame publisher
    \item Frame transformations publisher
    \item Model prediction
\end{itemize}

The target frame publisher node initialize a target frame having as reference the base of the robot. 

The frame transformation publisher node is in charge of transforming between different reference frames. In the simulation environment, the reference frame was the world frame, in the real robot it is the base of the robot. The Realsense sensor was added to the urdf file in order to use the sensor position and orientation for the ROS frame transformations.

The model prediction node obtains the observations and update the model for making a prediction for the next actions.

The ROS modules where setup to run at a speed of 50 Hz. The movement of the robot joints was controlled by relative angle and velocity, the angle rotation was fixed at 0.2 radians while the velocity was controlled by the predictions of the actions of the trained model.  

\section{Conclusion and Future Work}

In this work we presented a method for solving the freezing robot problem in a semi-static crowded and noisy environment. The method is composed of two separated modules. The "perception" module is in charge of detecting the target which in our case was either the hand or the arm of the person blocking the path of the robot. The "action" module is in charge of generating the gesture of the robot in order to touch the target. Having separated modules for the perception and the action of the robot, give the advantage of solve each problem separately. Each module can be replaced for other method of better accuracy. We used the Yolo v3 algorithm for fast detection, it could be replaced for other detection algorithms. For the action module, we used a policy trained on simulation using the PPO reinforcement learning algorithm. The generated policy was able to generate motion gestures that reached a dynamic target on simulation and tested on a real robot. Further work will measure the performance of the robot behaviour in real life-like scenarios. The code of the environment for training the policy in simulation can be found in: https://github.com/softbankrobotics-research/qi\_gym/tree/reaching\_target.

\section{Acknowledgements}
This work has been funded by the European Union's Horizon H2020 Research and Innovation program Crowdbot project under grand agreement No 779942.

\bibliographystyle{IEEEtran}
\bibliography{IEEEabrv,biblio}

\begin{thebibliography}{10}
\providecommand{\url}[1]{#1}
\csname url@samestyle\endcsname
\providecommand{\newblock}{\relax}
\providecommand{\bibinfo}[2]{#2}
\providecommand{\BIBentrySTDinterwordspacing}{\spaceskip=0pt\relax}
\providecommand{\BIBentryALTinterwordstretchfactor}{4}
\providecommand{\BIBentryALTinterwordspacing}{\spaceskip=\fontdimen2\font plus
\BIBentryALTinterwordstretchfactor\fontdimen3\font minus
  \fontdimen4\font\relax}
\providecommand{\BIBforeignlanguage}[2]{{%
\expandafter\ifx\csname l@#1\endcsname\relax
\typeout{** WARNING: IEEEtran.bst: No hyphenation pattern has been}%
\typeout{** loaded for the language `#1'. Using the pattern for}%
\typeout{** the default language instead.}%
\else
\language=\csname l@#1\endcsname
\fi
#2}}
\providecommand{\BIBdecl}{\relax}
\BIBdecl

\bibitem{joosse2017guide}
M.~Joosse and V.~Evers, ``A guide robot at the airport: First impressions,'' in
  \emph{Proceedings of the Companion of the 2017 ACM/IEEE International
  Conference on Human-Robot Interaction}, 2017, pp. 149--150.

\bibitem{sasaki2017long}
Y.~Sasaki and J.~Nitta, ``Long-term demonstration experiment of autonomous
  mobile robot in a science museum,'' in \emph{2017 IEEE International
  Symposium on Robotics and Intelligent Sensors (IRIS)}.\hskip 1em plus 0.5em
  minus 0.4em\relax IEEE, 2017, pp. 304--310.

\bibitem{mishraa2018issues}
N.~Mishraa, D.~Goyal, and A.~D. Sharma, ``Issues in existing robotic service in
  restaurants and hotels,'' in \emph{Proceedings of 3rd International
  Conference on Internet of Things and Connected Technologies (ICIoTCT)}, 2018,
  pp. 26--27.

\bibitem{tzafestas2018mobile}
S.~G. Tzafestas, ``Mobile robot control and navigation: A global overview,''
  \emph{Journal of Intelligent \& Robotic Systems}, vol.~91, no.~1, pp. 35--58,
  2018.

\bibitem{helbing1995social}
D.~Helbing and P.~Molnar, ``Social force model for pedestrian dynamics,''
  \emph{Physical review E}, vol.~51, no.~5, p. 4282, 1995.

\bibitem{zanlungo2011social}
F.~Zanlungo, T.~Ikeda, and T.~Kanda, ``Social force model with explicit
  collision prediction,'' \emph{EPL (Europhysics Letters)}, vol.~93, no.~6, p.
  68005, 2011.

\bibitem{ferrer2013robot}
G.~Ferrer, A.~Garrell, and A.~Sanfeliu, ``Robot companion: A social-force based
  approach with human awareness-navigation in crowded environments,'' in
  \emph{2013 IEEE/RSJ International Conference on Intelligent Robots and
  Systems}.\hskip 1em plus 0.5em minus 0.4em\relax IEEE, 2013, pp. 1688--1694.

\bibitem{vasquez2014inverse}
D.~Vasquez, B.~Okal, and K.~O. Arras, ``Inverse reinforcement learning
  algorithms and features for robot navigation in crowds: an experimental
  comparison,'' in \emph{2014 IEEE/RSJ International Conference on Intelligent
  Robots and Systems}.\hskip 1em plus 0.5em minus 0.4em\relax IEEE, 2014, pp.
  1341--1346.

\bibitem{chen2017socially}
Y.~F. Chen, M.~Everett, M.~Liu, and J.~P. How, ``Socially aware motion planning
  with deep reinforcement learning,'' in \emph{2017 IEEE/RSJ International
  Conference on Intelligent Robots and Systems (IROS)}.\hskip 1em plus 0.5em
  minus 0.4em\relax IEEE, 2017, pp. 1343--1350.

\bibitem{cruz2019enabling}
A.~Cruz-Maya, F.~Garcia, and A.~K. Pandey, ``Enabling socially competent
  navigation through incorporating hri,'' \emph{arXiv preprint
  arXiv:1904.09116}, 2019.

\bibitem{trautman2010unfreezing}
P.~Trautman and A.~Krause, ``Unfreezing the robot: Navigation in dense,
  interacting crowds,'' in \emph{2010 IEEE/RSJ International Conference on
  Intelligent Robots and Systems}.\hskip 1em plus 0.5em minus 0.4em\relax IEEE,
  2010, pp. 797--803.

\bibitem{fan2019getting}
T.~Fan, X.~Cheng, J.~Pan, P.~Long, W.~Liu, R.~Yang, and D.~Manocha, ``Getting
  robots unfrozen and unlost in dense pedestrian crowds,'' \emph{IEEE Robotics
  and Automation Letters}, vol.~4, no.~2, pp. 1178--1185, 2019.

\bibitem{trinh2015go}
T.~Q. Trinh, C.~Schroeter, J.~Kessler, and H.-M. Gross, ``“go ahead,
  please”: Recognition and resolution of conflict situations in narrow
  passages for polite mobile robot navigation,'' in \emph{International
  Conference on Social Robotics}.\hskip 1em plus 0.5em minus 0.4em\relax
  Springer, 2015, pp. 643--653.

\bibitem{vega2018planning}
A.~Vega, L.~J. Manso, R.~Cintas, and P.~N{\'u}{\~n}ez, ``Planning human-robot
  interaction for social navigation in crowded environments,'' in
  \emph{Workshop of Physical Agents}.\hskip 1em plus 0.5em minus 0.4em\relax
  Springer, 2018, pp. 195--208.

\bibitem{kamezaki2019preliminary}
M.~Kamezaki, A.~Kobayashi, Y.~Yokoyama, H.~Yanagawa, M.~Shrestha, and
  S.~Sugano, ``A preliminary study of interactive navigation framework with
  situation-adaptive multimodal inducement: Pass-by scenario,''
  \emph{International Journal of Social Robotics}, pp. 1--22, 2019.

\bibitem{buss2004introduction}
S.~R. Buss, ``Introduction to inverse kinematics with jacobian transpose,
  pseudoinverse and damped least squares methods,'' \emph{IEEE Journal of
  Robotics and Automation}, vol.~17, no. 1-19, p.~16, 2004.

\bibitem{guerrero1999neural}
A.~Guerrero-Gonzalez, J.~Lopez-Coronado, and F.~Garcia-Cordova, ``A neural
  network for target reaching with a robot arm using a stereohead,'' in
  \emph{IEEE SMC'99 Conference Proceedings. 1999 IEEE International Conference
  on Systems, Man, and Cybernetics (Cat. No. 99CH37028)}, vol.~2.\hskip 1em
  plus 0.5em minus 0.4em\relax IEEE, 1999, pp. 842--847.

\bibitem{moriarty1996evolving}
D.~E. Moriarty and R.~Miikkulainen, ``Evolving obstacle avoidance behavior in a
  robot arm,'' in \emph{From animals to animats 4: Proceedings of the fourth
  international conference on simulation of adaptive behavior}, vol.~4.\hskip
  1em plus 0.5em minus 0.4em\relax MIT Press, 1996, p. 468.

\bibitem{maass1997networks}
W.~Maass, ``Networks of spiking neurons: the third generation of neural network
  models,'' \emph{Neural networks}, vol.~10, no.~9, pp. 1659--1671, 1997.

\bibitem{haarnoja2018learning}
T.~Haarnoja, S.~Ha, A.~Zhou, J.~Tan, G.~Tucker, and S.~Levine, ``Learning to
  walk via deep reinforcement learning,'' \emph{arXiv preprint
  arXiv:1812.11103}, 2018.

\bibitem{peng2020learning}
X.~B. Peng, E.~Coumans, T.~Zhang, T.-W. Lee, J.~Tan, and S.~Levine, ``Learning
  agile robotic locomotion skills by imitating animals,'' \emph{arXiv preprint
  arXiv:2004.00784}, 2020.

\bibitem{quillen2018deep}
D.~Quillen, E.~Jang, O.~Nachum, C.~Finn, J.~Ibarz, and S.~Levine, ``Deep
  reinforcement learning for vision-based robotic grasping: A simulated
  comparative evaluation of off-policy methods,'' in \emph{2018 IEEE
  International Conference on Robotics and Automation (ICRA)}.\hskip 1em plus
  0.5em minus 0.4em\relax IEEE, 2018, pp. 6284--6291.

\bibitem{qureshi2016robot}
A.~H. Qureshi, Y.~Nakamura, Y.~Yoshikawa, and H.~Ishiguro, ``Robot gains social
  intelligence through multimodal deep reinforcement learning,'' in \emph{2016
  IEEE-RAS 16th International Conference on Humanoid Robots (Humanoids)}.\hskip
  1em plus 0.5em minus 0.4em\relax IEEE, 2016, pp. 745--751.

\bibitem{redmon2018yolov3}
J.~Redmon and A.~Farhadi, ``Yolov3: An incremental improvement,'' \emph{arXiv
  preprint arXiv:1804.02767}, 2018.

\bibitem{kuznetsova2018open}
A.~Kuznetsova, H.~Rom, N.~Alldrin, J.~Uijlings, I.~Krasin, J.~Pont-Tuset,
  S.~Kamali, S.~Popov, M.~Malloci, T.~Duerig \emph{et~al.}, ``The open images
  dataset v4: Unified image classification, object detection, and visual
  relationship detection at scale,'' \emph{arXiv preprint arXiv:1811.00982},
  2018.

\bibitem{schulman2017proximal}
J.~Schulman, F.~Wolski, P.~Dhariwal, A.~Radford, and O.~Klimov, ``Proximal
  policy optimization algorithms,'' \emph{arXiv preprint arXiv:1707.06347},
  2017.

\bibitem{busy2019qibullet}
M.~Busy and M.~Caniot, ``qibullet, a bullet-based simulator for the pepper and
  nao robots,'' \emph{arXiv preprint arXiv:1909.00779}, 2019.

\end{thebibliography}

\end{document}